# Backward Simulation in Bayesian Networks

**Robert Fung** and **Brendan Del Favero**
Institute for Decision Systems Research
4894 El Camino Real, Suite 110
Los Altos, CA 94022

## Abstract

Backward simulation is an approximate inference technique for Bayesian belief networks. It differs from existing simulation methods in that it starts simulation from the known evidence and works backward (i.e., contrary to the direction of the arcs). The technique's focus on the evidence leads to improved convergence in situations where the posterior beliefs are dominated by the evidence rather than by the prior probabilities. Since this class of situations is large, the technique may make practical the application of approximate inference in Bayesian belief networks to many real-world problems.

## 1 INTRODUCTION AND MOTIVATION

Because of its sound theoretical foundation in probability theory, the Bayesian belief network technology has become, in artificial intelligence, an important alternative architecture for reasoning to logic-based architectures (e.g., rule-based systems). Although efficient exact inference techniques (Shachter, 1986; Lauritzen, 1988, Shachter, 1990) for Bayesian belief networks can and have provided excellent solutions to many real-world problems, their applicability is limited, because exact inference is NP-complete (Cooper, 1990).

Because of this, significant research has been focused on finding efficient approximate inference methods. Most previous research has emphasized simulation methods (Pearl, 1987; Fung, 1989; Shachter, 1989; Chavez, 1990; Shwe, 1991a), which repeatedly draw sample values from the network's nodes using a sampling algorithm, and then use the relative frequencies of the sample values to estimate the probabilities of interest. Researchers try to find methods that converge quickly to the exact result, and to characterize the convergence properties of the simulation algorithms.

There are two basic classes of simulation methods: forward-simulation methods (Fung, 1989; Henrion, 1986; Shachter, 1989) and stochastic-simulation methods (Chavez, 1990; Pearl, 1987). Forward-simulation methods start each trial of the simulation by instantiating the source nodes (i.e., nodes with no predecessors) and then proceeding forward along the diagram arcs to instantiate each downstream node in turn. Because the sample values from one trial to the next are unrelated, the trials are independent. In stochastic-simulation methods, on the other hand, each trial begins by modifying the previous trial's instantiation. Each node's sample is chosen with respect to the current instantiations of neighboring nodes.

Because they are driven by the prior probabilities of upstream nodes, rather than by the likelihood of the observed evidence, forward-simulation methods converge slowly when faced with low-likelihood evidence (evidence that has low prior likelihood). Because of the way samples depend on the current instantiation, stochastic-simulation methods as a group are inefficient when there are deterministic or near-deterministic relationships in a network.

In this paper, we present the backward-simulation method for performing approximate probabilistic inference in Bayesian belief networks. Our method is closely related to forward-simulation methods, and is not susceptible to slow convergence in the presence of deterministic relationships as are stochastic simulation methods. In addition, the method is not as susceptible to slow convergence with low-likelihood evidence, the main problem with other methods of its class.

In Section 2, we present the notation used in this paper. In Section 3, we discuss forward simulation methods in detail. Section 4 provides the details on the backward-simulation method. In Section 5, we give a summary of the paper, and discuss directions for future research.

## 2 NOTATION

A Bayesian belief network is a directed acyclic graph $D$ with an associated probability distribution P. The set of nodes in a network is denoted by N. Individual nodes are denoted by capital letters, whereas general references to a node (such as "node $i$") are in lowercase italics. Each node $i$ in the network has a corresponding variable $X_i$ in P and a set of parents Pa($i$). If S is a set of nodes, then the set $X_S$ is the set of variables corresponding to the nodes in S.

The variable $X_i$ has a corresponding set of conditioning variables $X_{Pa(i)}$, called the parents of $X_i$. Each variable $X_i$ has a state space $\Omega_i$ and an associated probability distribution $P(X_i|X_{Pa(i)})$. The product of node probability distributions in a network is the joint probability distribution P of the variables associated with the graph $D$.

We represent evidence in Bayesian belief networks by setting the values of the appropriate nodes to their observed states. The set of nodes whose values have been observed is denoted by $N_e$. The nodes that are unobserved (i.e., that are in $N\backslash N_e$) are called state nodes.

The inference task for Bayesian belief networks is to compute answers to queries of the form $P(X_J|X_K)$, where all the observed evidence $N_e$ is typically a subset of the nodes in K. Many exact and approximate algorithms have been developed for addressing such queries.

Where there is little potential for confusion, the notation $P(A)$ will be used to mean $P(X_A)$.

## 3   IMPORTANCE SAMPLING

Importance sampling (Rubinstein, 1981) is a well-known technique for improving convergence in Monte Carlo simulation. It has been adapted for use in forward simulation models (Shachter, 1989). It provides the ability to instantiate the network from an arbitrary distribution $P_S$ instead of just from the joint distribution P as in logic sampling. To adjust for sampling from $P_S$ instead of from P, the weight Z associated with each trial is computed to be the ratio of the likelihood of the sample based on the network distribution to the likelihood of the sample based on the sampling distribution:

$$Z(X_N) = P(X_N) / P_S(X_N) \qquad (1)$$

The network probability $P(X_N)$ is always the product of node probabilities,

$$P(X_N) = \prod_{i \in N} P(x_i | x_{Pa(i)}) \qquad (2)$$

but the sampling distribution can be arbitrary. In the simplest form of the algorithm, the sampling distribution is the joint distribution over the unobserved nodes (the state nodes). In that case, where $P(X_{NS})=P_S(X_{NS})$, the ratio in Equation 1 is just $P(X_{N_e}|X_{NS})$, the overall likelihood of the evidence.

An informal argument for why this method works is as follows. If a large number of trials is done, the frequency of $X_i$ in the accumulated trials should be approximately $P_S(X_i)$. When multiplied by the weight we get the probability of $X_i$, $P(X_i)$. A formal proof of the convergence of this method is presented in (Geweke, 1989).

## 4   BACKWARD SIMULATION

The two defining features of backward simulation are the direction of simulation and the sampling method for drawing node values. The direction of sampling in the backward simulation method is outward from the evidence. In contrast, forward-simulation methods that work forward from the source nodes, whereas stochastic-simulation methods that have no particular direction of simulation. Like forward and stochastic simulation methods, backward simulation permits many possible orderings of the nodes to be sampled.

Backward simulation is a specialization of the importance-sampling inference method discussed in Section 3. Like forward simulation, backward simulation includes a node ordering step and a simulation step. The first step computes an ordering of network nodes, starting from the evidence nodes and working outward. In general, an ordering will contain only a subset of the nodes in the network. In the second step, the simulation trials are performed, with network nodes sampled in the predetermined order. A trial weight Z is computed for each trial, and is used to increment the counts of each distribution (i.e., query) of interest.

### 4.1   ORDERING

The three requirements for node ordering in backward simulation are flexible and allow significant variation:

1. A node must be instantiated before it is backward sampled,

2. A node's predecessors must be instantiated before the node is forward sampled, and

3. Each node in the network must be either (a) a node in the ordering or (b) a direct predecessor of a node in the ordering that is backward sampled.

Items 2 and 3(a) are the usual requirements for an ordering in forward simulation.

Because evidence nodes are instantiated, they can always be backward sampled. Leaf nodes also can always be backward sampled, because a dummy evidence node with uniform likelihoods can be attached to a leaf node without the posterior distribution being changed.

We shall use the simple network in Figure 1 to illustrate the ordering and sampling steps. Each node represents a probabilistic variable that is conditionally dependent on the nodes at the ends of the arrows pointing into it. Thus, the network in Figure 1 represents this probabilistic factorization:

$$P(ABCDE) = P(A) \, P(B|A) \, P(C|A) \, P(D|BC) \, P(E).$$

Node D is shaded to indicate that it is an evidence node.

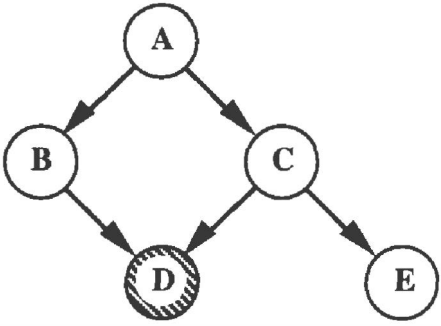

Figure 1: Simple Network.

Four different sampling orders for the example network shown in Figure 1 are {D,B,E}, {D,E,B}, {D,E,C}, and {D,C,E}. First, the evidence node D is sampled, instantiating nodes B and C. We then have two choices for instantiating node A, namely by backward sampling from either B or C. Finally, the value of E is determined by forward sampling from C.

The nodes in the sampling order will be denoted by $N_S$. This set is composed of the nodes to backward sampled, $N_b$, and the nodes to be forward sampled, $N_f$. In this example, if we take $N_S$ to be {D,B,E}, then $N_b$ is {D,B} and $N_f$ is {E}.

## 4.2 SAMPLING

Two sampling methods are used in backward simulation: forward sampling and backward sampling. In forward sampling, a node's distribution is sampled only after all the node's predecessors have been sampled. Forward sampling a node sets the value for the node itself. The probability distribution on which the random sampling is based is determined by the values of the node's predecessors.

Backward sampling differs from forward sampling in each of these three aspects. First, sampling from a node's probability distribution occurs only after the node itself has been instantiated. Second, sampling from a node's probability distribution does not determine the node's value but rather the values of the node's predecessors. Third, the sampling is based not on a particular conditional probability distribution but rather on normalized likelihoods of the node's conditional probability distribution.

Which of the two methods is used to sample a node depends on the network topology. For nodes with an evidence node as a descendent, sampling from a node's conditional probability occurs after the node has been instantiated, and it determines the values for the node's predecessors. For nodes with no downstream evidence nodes, forward sampling is used. Any of the algorithms developed for forward simulation can be used.

We would like to emphasize the distinction between sampling (a selection from a set of choices) and simulation (the process driving the sampling). Indeed, in forward simulation, they are linked: forward simulation involves only forward sampling. However, the process of backward simulation does involve both backward and forward sampling.

### 4.2.1 Backward Sampling

Backward sampling from a node's probability distribution instantiates those of the node's predecessors that are not already instantiated. We denote the instantiated value of node $i$ by $x_i$, the parents of $i$ that are uninstantiated by $Pa^u(i)$ and the instantiated parents by $Pa^*(i)$.

The backward-sampling procedure instantiates the uninstantiated parent nodes according to the following probability distribution:

$$P_s(Pa^u(i)) = \frac{P(x_i | X_{Pa^u(i)}, x_{Pa^*(i)})}{\text{Norm}(i)}, i \in N_b. \quad (3)$$

The numerator of the preceding expression is the likelihood of the current state of node $i$ given a particular state of the parents of node $i$. The denominator, Norm($i$), is a normalization constant that ensures that the terms in this distribution sum to 1. Norm($i$) is computed as follows:

$$\text{Norm}(i) = \sum_{y \in XP(Pa^u(i))} P(x_i | y, x_{Pa^*(i)}).$$

The set $XP(Pa^u(i))$ is the set of all possible conditioning cases of the uninstantiated parents of node $i$. For nodes with $n$ binary-valued uninstantiated parents, this set is of size $2^n$.

Normalization constants can be precomputed if there is a fixed order, or they can be cached as computation proceeds.

### 4.2.2 Forward Sampling

In forward sampling, the sampling distribution for a node is the same as the node's probability distribution:

$$P_s(i) = P(x_i | x_{Pa(i)}), \ i \in N_f. \quad (4)$$

### 4.3 SCORING

After all of the nodes have been instantiated, the weight for each trial can therefore be computed by combining Equations 1 through 4:

$$Z(x) = \frac{\prod_{i \in N} P(x_i | x_{Pa(i)})}{\left(\prod_{j \in N_b} \frac{P(x_j | x_{Pa(j)})}{\text{Norm}(j)}\right)\left(\prod_{j \in N_f} P(x_j | x_{Pa(j)})\right)}$$

This can be simplified to

$$Z(x) = \prod_{i \in N \setminus N_s} P(x_i \mid x_{Pa(i)}) \prod_{j \in N_b} \text{Norm}(j) \quad (5)$$

### 4.4 EXAMPLE

Let us step through an example of backward simulation. Consider the five-node network in Figure 1. All of the nodes are binary-valued: the values for A, for instance, are taken to be $a_1$ and $a_2$, whereas the possible joint values (or states) of nodes B and C are $b_1c_1$, $b_1c_2$, $b_2c_1$, and $b_2c_2$. The value of the evidence node D is observed to be $d_2$. Suppose that the sampling order {D,B,E} is used.

Table 1 is the conditional probability table for node D. For instance, $P(D=d_2 \mid B=b_1, C=c_2) = p_{122}$.

Table 1: Probability distribution for D given B and C

| P( D \| BC ) | $b_1c_1$ | $b_1c_2$ | $b_2c_1$ | $b_2c_2$ |
|---|---|---|---|---|
| $d_1$ | $p_{111}$ | $p_{121}$ | $p_{211}$ | $p_{221}$ |
| $d_2$ | $p_{112}$ | $p_{122}$ | $p_{212}$ | $p_{222}$ |

Since D is observed to be $d_2$, the sampling distribution for the parents of D is based on the second row of Table 1. The sampling distribution, over the states in $XP(Pa^u(D))$, $\{b_1c_1, b_1c_2, b_2c_1, b_2c_2\}$, is shown in Table 2.

Table 2: Sampling Distribution for B and C

| $P_S$( BC \| D=$d_2$ ) | $b_1c_1$ | $b_1c_2$ | $b_2c_1$ | $b_2c_2$ |
|---|---|---|---|---|
| | $\frac{p_{112}}{\alpha}$ | $\frac{p_{122}}{\alpha}$ | $\frac{p_{212}}{\alpha}$ | $\frac{p_{222}}{\alpha}$ |

The constant $\alpha$ normalizes the terms to sum to 1, namely:

$$\alpha = \text{Norm}(D=d_2) = p_{112} + p_{122} + p_{212} + p_{222}.$$

Suppose that the sampling step chooses joint state $b_1c_2$ and sets the states of B and C to these values.

Next, we would sample node B to set node A to one of the states $\{a_1, a_2\}$ according to the distribution over these states. Table 3 shows the conditional probability distribution of B given A, whereas Table 4 shows the sampling distribution for A.

Table 3: Probability of B given A

| P( B \| A ) | $a_1$ | $a_2$ |
|---|---|---|
| $b_1$ | $q_{11}$ | $q_{21}$ |
| $b_2$ | $q_{12}$ | $q_{22}$ |

Table 4: Backward Sampling Distribution for A

| $P_S$( A \| B=$b_1$ ) | $a_1$ | $a_2$ |
|---|---|---|
| | $\frac{q_{11}}{\beta}$ | $\frac{q_{21}}{\beta}$ |

As before, the constant $\beta$ normalizes the terms:

$$\beta = \text{Norm}(B=b_1) = q_{11} + q_{21}.$$

Suppose that the sampling step selects state $a_2$.

Finally, we would use forward sampling to set node E to one of the states ($e_1$, $e_2$) according to the distribution given in Table 6, which is identical to the second column of Table 5.

Table 5: Probability of E given C

| P( E \| C ) | $c_1$ | $c_2$ |
|---|---|---|
| $e_1$ | $r_{11}$ | $r_{21}$ |
| $e_2$ | $r_{12}$ | $r_{22}$ |

Table 6: Sampling Distribution for E

| $P_S$( E \| C=$c_2$ ) | $e_1$ | $e_2$ |
|---|---|---|
| | $r_{21}$ | $r_{22}$ |

In forward sampling, the sampling distribution is the same as the probability distribution, so no normalization constant is necessary (the terms already sum to 1). Suppose that the sampling step sets E to $e_1$.

This example trial has instantiated the network to the joint state $a_2b_1c_2d_2e_1$. The trial score that would be added to the beliefs of the currently-selected states of each node would be

$$Z = P(A=a_2)P(C=c_2|A=a_2)\text{Norm}(D=d_2)\text{Norm}(B=b_1).$$

### 4.5 CORRECTNESS AND CONVERGENCE

Backward simulation meets the single constraint for a valid importance sampling procedure: no point in the joint state space of the prior distribution with positive probability can have a probability of zero in the sampling distribution.

As a form of importance sampling with likelihood weighting, backward simulation inherits the convergence properties of importance sampling (Shachter, 1989). That is, the beliefs generated by the simulation are guaranteed to converge to their true values, with the errors decreasing in proportion to the square root of the number of trials.

## 5 DISCUSSION

### 5.1 BENEFITS AND COSTS

The backward-simulation method is well-suited to inference in situations with low-likelihood evidence. By working from the evidence, the method focuses on instantiating those scenarios that are most compatible with the observed network state, rather than with the prior distribution. The effect of the prior distribution is taken into account by the trial weights. If the prior probabilities are diffuse, compared to the evidence likelihoods, the weights also will be diffuse, and the inference method will converge to the correct solution much faster than would forward-simulation methods.

This benefit of backward sampling over forward sampling for low-likelihood evidence is illustrated by the two-node network in Figure 2. Node T has been observed at value $t_1$, and node S has two states, $s_1$ and $s_2$.

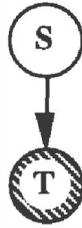

Figure 2: Two-node Network With Low-likelihood Evidence.

Suppose that the prior and conditional probabilities are as follows, with $0 < \varepsilon \ll \delta \ll 1$:

$$P(S = s_1) = \delta$$
$$P(S = s_2) = (1 - \delta)$$
$$P(T = t_1 | S = s_1) = (1 - \varepsilon)$$
$$P(T = t_1 | S = s_2) = \varepsilon$$

Using exact inference (Bayes's rule), we can show that, although the prior for state $s_1$ is much less than that for state $s_2$, $s_1$ is the most likely explanation for the evidence:

$$P(s_1 | t_1) = \frac{P(s_1 t_1)}{P(s_1 t_1) + P(s_2 t_1)}$$
$$= \frac{\delta(1 - \varepsilon)}{\delta(1 - \varepsilon) + (1 - \delta)\varepsilon}$$
$$\approx \frac{\delta}{\delta + \varepsilon} = 1 - \frac{\varepsilon}{\delta + \varepsilon} \approx 1$$
$$P(s_2 | t_1) \approx \frac{\varepsilon}{\delta + \varepsilon} \approx 0$$

Table 7 shows the forward sampling distribution for S. If we use forward sampling from S to T, most trials will set S to $s_2$, and would set the trial score Z to $\varepsilon$, a relatively small number. The belief corresponding to $s_2$ will be augmented by this score, whereas the belief of $s_1$ will remain at zero. After many trials (on average, about $1/\delta$), node S will be set to $s_1$ and the trial score Z would be $1-\varepsilon$. Now, because of this one trial, the belief of $s_1$ will be much greater than the belief of $s_2$; we will have discovered, after much work, that $s_1$ is the most likely explanation for the evidence observed.

Table 7: Forward Sampling Distribution for S

| P(S) | $s_1$ | $s_2$ |
|---|---|---|
| | $\delta$ | $1-\delta$ |

Table 8 shows the backward sampling distribution for S. If we use backward sampling from T to S, S will be set to state $s_1$ in the preponderance of trials. The belief of $s_1$ will be augmented in each trial by $Z=\delta$, and the belief of $s_2$ by zero. Thus, from the start, the simulation is more in line with the most probable diagnosis, $s_1$.

Table 8: Backward Sampling Distribution for S

| $P_S(S | T=t_1)$ | $s_1$ | $s_2$ |
|---|---|---|
| | $1-\varepsilon$ | $\varepsilon$ |

The main cost of backward sampling is the computational resources required for computing the normalization constants. Although in general the costs grow exponentially with the number of predecessors, the costs can be reduced where there are special network structures such as invertible continuous functions or noisy-or relationships

Backward simulation is related to the method of evidential integration (Chin, 1987; Fung, 1989) that has been suggested for use with simulation methods. In evidential integration, arc reversals are used as a pre-processing step to integrate the evidence into the network, to convert extremal likelihoods to less extreme likelihoods. Evidence integration is computationally expensive; backward simulation does part of what evidence integration does (when it computes the normalization constants), at a fraction of the cost.

For networks in which the conditional probabilities do not change, the normalization constants can be precomputed and cached, taking much of the work out of each trial.

### 5.2 EXPERIMENTS

We have run some preliminary experiments comparing the performance of forward simulation with backward simulation. They were based on the network in Figure 1, as described in (Fung, 1989). The probabilities are given in Table 9. D is observed in state $d_2$ and E is observed in state $e_1$.

The test routines were written in Macintosh Common Lisp. We tested a forward simulation method against the backward simulation method. Both use likelihood weighted scoring.

Table 9: Probabilities for the Experimental Network

| | | | | |
|---|---|---|---|---|
| P(A) | $P(a_1)$ | = 0.20 | | |
| P(B\|A) | $P(b_1\|a_1)$ | = 0.80 | $P(b_1\|a_2)$ | = 0.20 |
| P(C\|A) | $P(c_1\|a_1)$ | = 0.20 | $P(c_1\|a_2)$ | = 0.05 |
| P(D\|BC) | $P(d_1\|b_1c_1)$ | = 0.80 | $P(d_1\|b_2c_1)$ | = 0.80 |
| | $P(d_1\|b_1c_2)$ | = 0.80 | $P(d_1\|b_2c_2)$ | = 0.05 |
| P(E\|C) | $P(d_1\|b_1c_1)$ | = 0.80 | $P(d_1\|b_1c_1)$ | = 0.60 |

As was done in the previously cited paper, we perform a large number (250) of runs. In each run, we measure the accuracy of each simulation method at 100, 200, 500, 1000, and 2000 trials, using the absolute-value error function below:

$$Error(t) = \frac{1}{N} \sum_{k=1}^{N} \sum_{i \in N} \sum_{j=1}^{2} \left| B_k(x_{ij}, t) - P(x_{ij}) \right|.$$

Here, $B_k(x_{ij}, t)$ is the belief (probability estimate) for state $j$ of node $i$ on trial $t$ of the $k$th run. The error, averaged over all runs, is presented in Figure 3. The standard deviation of the errors in the runs is presented in Figure 4.

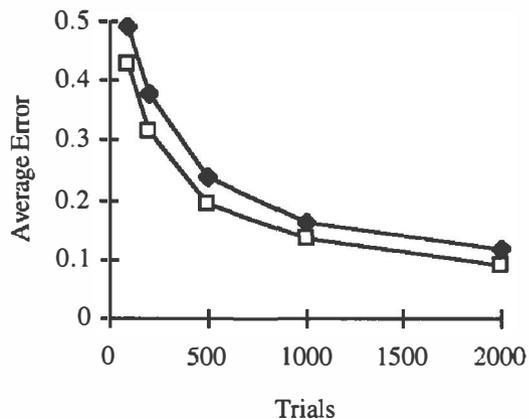

Figure 3: Error vs. Trials for Two Simulation Methods: Forward (Diamonds) and Backward (Squares)

Figures 3 and 4 show that backward and forward simulation are performing equivalently well, with the backward simulation method showing slightly lower average error values. This is to be expected: given a large enough number of trials, both methods will converge to the same answer. In this example, the evidence is not particularly unlikely, thus there is no particular benefit for backward sampling. Most importantly, Figures 3 and 4 show that backward simulation works as an inference method.

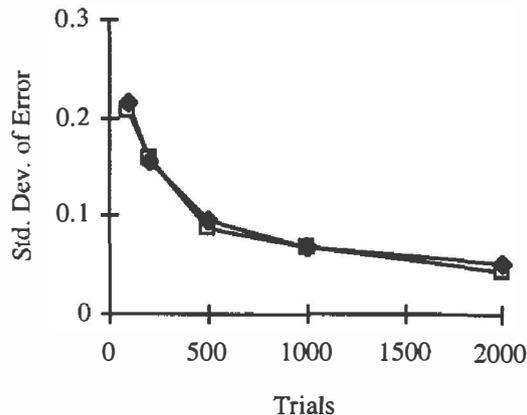

Figure 4: Standard Deviation vs. Trials: Forward (Diamonds) and Backward (Squares)

Although both simulation methods will converge to the same answer, they may do so at different speeds. This is the motivation of the next set of runs, which were performed on the same network with the same distributions, with the exception of the modified entries listed in Table 10. The evidence in these runs is that node D is set to $d_1$. These modifications make the observed evidence much less likely than before.

Table 10: Modifications to Probabilities in Table 9 for Extreme-probability Experiment

| | | | | |
|---|---|---|---|---|
| P(D\|BC) | $P(d_1\|b_1c_1)$ | = 0.001 | $P(d_1\|b_2c_1)$ | = 0.0001 |
| | $P(d_1\|b_1c_2)$ | = 0.0001 | $P(d_1\|b_2c_2)$ | = 0.05 |

We record the performance of each method at 10, 20, 50, 100, and 200 trials. The average error and standard deviation are presented in Figures 5 and 6.

Figures 5 and 6 show that at low trial numbers, backward simulation is consistently closer to the true probability than forward simulation.

The conclusion is that we can expect good performance of backward simulation at a low numbers of trials in networks with low-likelihood evidence. Of course, much more work is necessary to characterize and understand the conditions under which each simulation method performs better than the other. There are cases in which backward simulation would perform worse than forward simulation: backward simulation is subject to the proof of Dagum (1993) that, in the worst case, all approximate probabilistic methods are NP-hard.

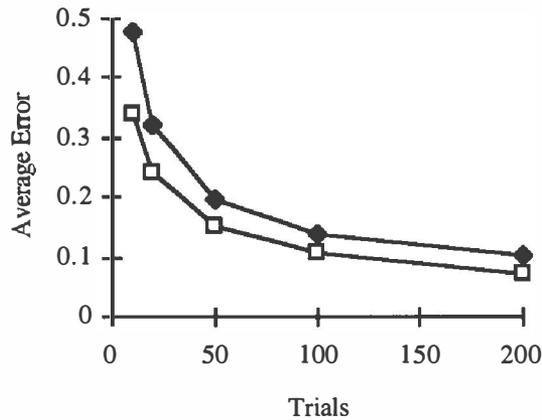

Figure 5: Error vs. Trials for Second Experimental Run: Forward (Diamonds) and Backward (Squares)

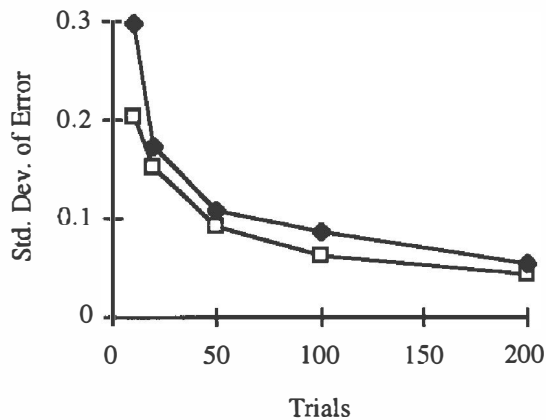

Figure 6: Standard Deviation vs. Trials for Second Experimental Run: Forward (Diamonds) and Backward (Squares)

### 5.3 THE SIGNIFICANCE OF NODE ORDERING

Backward simulation provides a new flexibility in devising strategies for instantiating the network during a simulation trial. One strategy would be to use backward sampling wherever possible. Another would use forward sampling everywhere (this is just forward simulation). In between these extremes, there are many possibilities for mixing both forward and backward sampling.

There are often multiple possible node ordering in forward simulation as well. However, node ordering has no impact on the sampling distribution for a particular node: the distribution is always based on the node's predecessors, which are always instantiated before the node itself is.

In backward simulation, on the other hand, the sampling order can affect the sampling distribution. This may have a significance in convergence properties of the simulation. For instance, in the example of Section 4.4, two possible sampling orders are {D,E,B} and {D,E,C}. Using the former, A is backward sampled from B, whereas using the latter, A is backward sampled from C. Depending on the structure of the joint distribution P(ABC), the sampling distribution for A could be quite different between the two orders. One ordering may be better than the other in terms of simulation performance.

The ordering {A,B,C,E} presents a different type of flexibility, or ambiguity: it is valid under this order to instantiate C by forward sampling from A or by backward sampling from D. There may be a difference in simulation performance under the different interpretations of this ordering.

The added flexibility is exciting, in that it provides an opportunity to develop heuristics governing when to sample a node backward or forward. The hope is that with backward and forward simulation in the probabilistic tool chest, as well as evidence integration and other approximate methods, a simulation run can be optimized based on the particulars of the network being considered.

## 6 FUTURE RESEARCH AND APPLICATIONS

There are many interesting avenues of research for backward simulation. Dynamic node ordering (i.e., changing $N_S$ within a single trial or between trials) is possible and may provide a way to improve performance. Also, it is possible to group nodes for sampling. For example, in Figure 1, B and C could be aggregated together for the purpose of sampling the value of A.

One of the most promising areas of research is the combination of the backward-simulation method with its dual, forward simulation. The combination would provide a complete probabilistic architecture that allows both data-driven and causal reasoning. Such an architecture might have the promise of attacking such problems as natural-language understanding or speech recognition. We are currently testing this architecture on two-level networks with noisy-or relationships between the nodes such as the QMR-DT network (Shwe, 1991b).

This combination may make feasible the application of Bayesian belief networks to many real-world settings that current techniques cannot handle. For example, it should be applicable to situations that have large and dynamic state spaces, and strong evidence. Many sensory situations (e.g., vision) seem to have this flavor. It seems to match well with how people reason under uncertainty — by reasoning from evidence to conclusions. This has the promise of making explanation of the results of backward simulation more intuitive than for exact inference methods.

234     Fung and Del Favero

**Acknowledgments**

This work was supported in part by NSF Grant IRI-9120330.

This work has benefited greatly from discussions with Mark Peot and Kuo-Chu Chang and from the comments of the reviewers of this paper.